%% file: main.tex
\pdfoutput=1
\documentclass[12pt]{article}

\usepackage[margin=1in]{geometry}
\usepackage{setspace}
\usepackage[T1]{fontenc}
\usepackage[utf8]{inputenc}
\usepackage{mathptmx}
\usepackage[scaled=0.92]{helvet}
\usepackage{courier}
\usepackage{microtype}

\usepackage{amsmath}
\usepackage{graphicx}
\usepackage{booktabs}
\usepackage{enumitem}
\usepackage[font=small,labelfont=bf,skip=6pt]{caption}
\usepackage{placeins}
\usepackage{xcolor}
\definecolor{linkblue}{rgb}{0.00,0.25,0.45}

\usepackage[round]{natbib}
\usepackage{authblk}

\usepackage{xurl}
\usepackage[colorlinks=true,linkcolor=linkblue,citecolor=linkblue,urlcolor=linkblue]{hyperref}
\hypersetup{pdftitle={Beyond Moral Performance: Empirical Assessment of LLMs' Moral Competence Through Judgment Stability}}

\title{Beyond Moral Performance: Empirical Assessment of LLMs' Moral Competence Through Judgment Stability}

\author[1]{Eoin O'Doherty\thanks{Equal contribution.}}
\author[1]{Nicole Weinrauch$^*$}
\author[1]{Andrew Talone$^*$}
\author[1]{Uri Klempner$^*$}
\author[2]{Pablo Biedma$^*$}
\author[3]{Linus Ta-Lun Huang$^*$}
\author[4]{Xiaoyuan Yi\thanks{Corresponding author. Email: xiaoyuanyi@microsoft.com}}
\author[4]{Xing Xie}
\author[1,5]{Yi Zeng}

\affil[1]{Schwarzman College, Tsinghua University, Beijing, China}
\affil[2]{Department of Computer Science and Technology, Tsinghua University, Beijing, China}
\affil[3]{The Chinese University of Hong Kong, Hong Kong, Hong Kong S.A.R.}
\affil[4]{Microsoft Research Asia, Beijing, China}
\affil[5]{Chinese Academy of Sciences, Beijing, China}

\date{}

\begin{document}
\maketitle

\begin{abstract}
Evaluations of large language models (LLMs) have largely measured moral performance, whether their outputs are morally appropriate, rather than moral competence, the capacity to judge on the basis of morally relevant considerations. We test for competence by varying morally irrelevant features of moral dilemmas. Because moral dilemmas rarely admit a single correct answer, we assess the stability of framework preferences, a model's relative ordering of five normative ethical frameworks, rather than of verdicts. We test 16 LLM configurations, reasoning and non-reasoning, on 18 dilemmas in six thematic domains for reliability across repeated runs and stability across prompt length, thematic domain, and reporting format. The configurations share broadly the same preference ordering: libertarianism falls far below the other frameworks in all 16, and care and virtue ethics occupy the top two positions in 15. This ordering is highly reliable across runs and largely preserved across reporting formats, and we find no evidence that it depends on thematic domain. Prompt length, however, affects it: longer prompts lower the scores of libertarian, virtue ethics, and care ethics outcomes but not those of utilitarian and deontological outcomes, so that with long prompts care ethics, utilitarianism, and virtue ethics can no longer be distinguished, while libertarianism remains last. Reasoning configurations show closer agreement between reporting formats but less reliability in rankings. Since human moral judgement also shows sensitivity to morally irrelevant features, perfect stability cannot be the criterion for competence; on balance, the pattern of our data remains more compatible with the existence of moral competence than with its absence, with the sensitivity to prompt length as the main qualification.
\end{abstract}


\input{sec_intro}
\input{sec_related}
\input{sec_method}
\input{sec_result}
\input{sec_discuss}
\input{sec_conclu}
\input{sec_statements}

\bibliographystyle{chicago-bmp}
\bibliography{references}

\clearpage
\input{appendix}

\end{document}

%% file: sec_intro.tex
\section{Introduction}\label{sec:intro}

Large language models \citep[{LLMs;}][]{team2023gemini,guo2025deepseek_r1,openai2024gpt4o} exhibit moral reasoning capabilities \citep{dillion2025ai,jiang2021can}. As they are deployed across social domains \citep{horton2023large,pataranutaporn2025simulating} and increasingly act on behalf of their users \citep{andersson2025companionship,thirunavukarasu2023large,zhang2025rise}, they will shape decision-making in morally significant situations \citep{chiu2024dailydilemmas,sandmann2025benchmark}. It is therefore essential to assess whether LLMs have the practical reasoning competence that moral judgement requires, and not only the theoretical reasoning competence they display in domains such as mathematics \citep{cheung2025large,christian2020alignment,dung2023current,ji2025moralbench}. The two kinds of competence cannot be assessed in the same way. A mathematical answer can be scored against a known solution; a moral judgement often cannot.

Following \citet{haas2026roadmap}, we distinguish moral performance, the production of morally appropriate outputs, from moral competence, the capacity to produce those outputs on the basis of morally relevant considerations and principles. Ideally, a system's computational process should structurally match the problem it solves, that is, reflect the normatively correct reasoning procedure for solving it, which also ensures reliable generalisation to other questions in the same domain \citep{haas2026roadmap}. However, because LLMs function by predicting the next probable token based on statistical distributions, rather than employing symbolic reasoning processes, we cannot easily determine whether they use the appropriate reasoning procedure or a superficial ``facsimile''---such as memorized answers from the training data or simple heuristics \citep{zaman2025chain,haas2026roadmap,song2026large}. This is the ``facsimile problem''.

Because LLMs have an enormous capacity to produce superficially good outputs, assessing moral competence requires tests designed to pull the two possibilities apart. Without reliable tools for inspecting internal processes, systematic breakdowns in performance remain one of the strongest sources of evidence for a lack of underlying competence. The converse inference is weaker: stability has many possible producers, only some of which involve competence, so a system that does not break down provides only weak evidence that it is competent.

Stable judgement across morally irrelevant changes in context is accordingly a key indicator of moral competence \citep{bonagiri-etal-2024-sage,oh2025robustness,zhou2024unveiling}, and it also supports reproducibility and fairness in decision-support applications \citep{cheung2025large,krugel2023chatgpt}. The competence and facsimile accounts predict different patterns of stability. If a model's judgements are produced by moral competence, they should survive variation along every morally irrelevant dimension. If they are produced by a facsimile, they may survive variation along one dimension while failing along another. Stability across several independent dimensions is therefore a more demanding test than stability along any one of them. Assessing stability poses a further challenge, however, because there is rarely just one right answer in moral contexts and moral competence involves context-sensitive trade-offs among competing considerations \citep{sorensen2024position}. We do not ask, then, whether an LLM consistently produces a single correct answer. We ask whether its framework preferences, the relative ordering it gives to five normative ethical frameworks, remain stable. High stability of framework preferences is consistent with moral competence \citep{jiao2025llm,scherrer2023evaluating}.

We address three research questions:

\begin{description}[leftmargin=2.2em, labelindent=0pt, itemsep=2pt, topsep=4pt]
  \item[RQ1:] Which of five normative ethical frameworks (utilitarianism, deontology, virtue ethics, care ethics, and libertarianism) do state-of-the-art LLMs prefer when faced with structured moral questions?
  \item[RQ2:] Do the framework preferences of LLMs show the reliability and stability that moral competence entails? Specifically, we test whether they are reliable across repeated runs and stable across three dimensions of morally irrelevant variation: prompt length (short versus extended scenarios), reporting format (morality scores versus rankings), and thematic domain (dilemmas from different areas of ethical life).
  \item[RQ3:] Is reasoning before answering, as reasoning models such as DeepSeek-R1 do, associated with greater reliability and stability?
\end{description}

Our contribution is twofold. (1) We offer a way to test for moral competence when moral dilemmas lack a unique correct answer: we assess the stability of framework preferences, rather than of verdicts, along several independent dimensions of morally irrelevant variation. (2) Applying this test to 16 LLM configurations, we find that they share broadly the same preference ordering, which is highly reliable and largely preserved across reporting formats but sensitive to prompt length: longer prompts lower the scores of some frameworks more than others and change the ordering among the top three frameworks; read in light of the asymmetry between instability and stability, this pattern is, on balance, more compatible with moral competence than with its absence, with the sensitivity to prompt length as the main qualification.

Section~\ref{sec:related} situates our approach in work on evaluating the moral judgements of LLMs. Section~\ref{sec:methods} sets out the competence--performance distinction and the study design, and Section~\ref{sec:results} reports the results. Section~\ref{sec:discussion} discusses what they imply for moral competence, for the role of reasoning, and for the preference ordering the configurations share, and states the limitations of the study.

%% file: sec_related.tex
\section{Related Work}\label{sec:related}

In this paper, we do not try to settle the question of whether AI systems can be moral agents \citep{brozek2019can}. Our concern is narrower and more empirical. Contemporary LLMs exhibit identifiable moral orientations \citep{rozado2023political,simmons-2023-moral}, and we ask how those orientations should be evaluated. By moral orientations we mean the values a model expresses. Work on the evaluation question has moved through three phases, each tightening what counts as evidence of a capacity for moral reasoning.

The initial question was purely descriptive: if LLMs exhibit moral orientations in their output, which ones? Early work answered it by borrowing survey instruments from empirical moral psychology and administering them to LLMs as if to human survey respondents. \citet{abdulhai2022moral}, for example, applied the Moral Foundations Questionnaire \citep[{MFQ;}][]{graham2011mapping} to GPT-3 models and showed that their moral judgements were more similar to those of conservative human respondents. \citet{arora2023probing} used the World Values Survey \citep{Haerpfer2022} and Hofstede's survey instruments based on Cultural Dimensions Theory \citep{hofstede2001culture}. They showed that multilingual models did pick up differentiated values across the 13 cultures tested, but that those differences correlated only weakly with the cross-cultural survey results from the respective populations. \citet{fraser2022does} drew on normative moral frameworks (utilitarianism, deontology, etc.), in addition to empirical ones, to show that Delphi \citep{jiang2021can}, a model specifically trained to perform moral reasoning, largely reflected the moral leanings of its annotators. Subsequent work built a generation of large benchmarks, either by adapting existing instruments and sources \citep{nie2023moca,ren-etal-2024-valuebench,sachdeva2025normative,scherrer2023evaluating,zhao2024worldvaluesbench} or by synthesising new items with LLMs specifically for testing LLMs \citep{biedma2024humannormsunveilingunique,duan2024denevil,samway-etal-2025-language}. Taken together, this literature established that LLMs have measurable moral orientations, as expressed in their responses to survey-like questions designed to assess moral values. What it left open was whether those orientations extend beyond this specific domain of responses.

This concern pushed the field towards a different target: not what models explicitly say they value, but the underlying moral capabilities expressed in their broader behaviours \citep{emelin2021moral,hendrycks2020aligning}. \citet{jin2022make} tested whether LLMs could identify cases in which rule violations are morally permissible, as humans do. \citet{sorensen2024value} assessed whether models could identify the plural values at stake in a given situation, and proposed Kaleido, a fine-tuned model that classifies how relevant each value is to a situation and whether it supports or opposes an action. \citet{chiu2024dailydilemmas} used dilemmas from everyday contexts, such as family, workplace, and friendships, to evaluate whether LLMs produce responses in line with human values. Most recently, \citet{chiu2025morebench} moved beyond final outputs to examine reasoning traces (i.e., the text an LLM writes out before it delivers its final answer), evaluating whether (part of) the procedure by which a model arrives at a judgement tracks the procedure a human reasoner would use. Each of these tests, however, still scores outputs or traces against what humans judge appropriate, and a model relying on a facsimile of moral reasoning could pass them.

The facsimile problem (Section~\ref{sec:intro}) has accordingly motivated a shift in evaluative strategy. If we cannot inspect the entire reasoning process directly, we can at least ask whether the outputs behave as the results of genuine reasoning would. Among the behavioural signatures proposed, stability under morally irrelevant perturbation has emerged as central \citep{atil2024non,haas2026roadmap,jang-lukasiewicz-2023-consistency,jiao2025llm}. Stability has also been studied as a dimension of LLM reasoning competence more broadly, in domains including values \citep{du2025investigating,moore2024large}, personality \citep{tosato2026persistent}, legal reasoning \citep{blair2025llms}, mathematical and logical reasoning \citep{liu2024aligning}, and code generation \citep{min2023beyond}.

Within the moral domain, existing work provides evidence that LLM judgements can be unstable. \citet{bonagiri-etal-2024-sage} showed that LLMs can produce unstable judgements across semantically equivalent moral questions, and \citet{oh2025robustness} demonstrated substantial sensitivity to seemingly superficial changes in prompt formulation. \citet{zhou2024unveiling} further showed that stability in moral judgement is lower in scenarios with greater moral ambiguity, where no option is clearly morally superior. This last finding exposes a methodological problem. When a dilemma admits several defensible verdicts, instability at the level of the verdict runs together (a) a failure to track morally relevant considerations and (b) indifference between verdicts that are equally defensible. Only (a) counts against moral competence. If there is no uniquely correct verdict, stability cannot simply mean reproducing the same answer across equivalent prompts. This motivates our approach of evaluating the stability of a model's framework preferences along several independent dimensions of morally irrelevant variation at once.

%% file: sec_method.tex
\section{Methods}\label{sec:methods}

The study elicits moral judgements from each LLM in a structured task. A model is given a dilemma together with five possible outcomes, each recommending a course of action and justifying it by the characteristic consideration of one normative ethical framework: aggregate welfare for utilitarianism, duty and human dignity for deontology, character for virtue ethics, relational responsibility for care ethics, and non-interference for libertarianism. The model evaluates the five outcomes in two separate task: in one, it assigns each a morality score from 0 to 100; in the other, it ranks them from most to least moral. From these evaluations we recover the model's framework preferences, the order in which it places the five frameworks. Each evaluation is repeated in 10 runs, and the dilemmas are presented at three lengths and drawn from six thematic domains. The test is whether a model's framework preferences are reliable across runs and stable across variations that should not change a moral judgement.

\subsection{Testing Competence Through Stability}\label{sec:stability-test}

Our methodology relies on a distinction, adapted from linguistics \citep{chomsky2014aspects}, between performance and competence.\footnote{Chomsky's original distinction presupposed an internal grammar whose structure could in principle be characterised independently of its implementation in production. Whether a comparable structure exists for moral reasoning, in humans or in LLMs, is a substantive question we do not try to settle. Readers who reject the analogy can, without loss, read our argument as concerning whatever capacity-level property would explain stable, principled moral judgement in a system.} In linguistics, competence is a speaker's knowledge of their language, and performance is the actual use of that knowledge in producing and understanding sentences, which is also shaped by factors such as memory limitations, distraction, and error. Occasional slips do not show that a speaker lacks the grammar; errors that vary systematically with features the grammar treats as irrelevant are better evidence of a gap. The moral analogue, which we take from \citet{haas2026roadmap} distinguishes the moral judgements a system actually produces in a given context from the underlying capacity to produce those judgements on the basis of morally relevant considerations. Because only performance is observable, the methodological question is how to test for competence through performance. Our answer rests on a three-step inference. (1) If a model's judgements are produced by moral competence, its framework preferences should remain largely stable when morally irrelevant features of the prompt are varied. (2) Systematic instability under such variation is therefore evidence against competence: if a model's framework preferences shift when irrelevant features are perturbed, that is good evidence that they were not produced by tracking the morally relevant features in the first place. (3) Stability, by contrast, is only weak evidence for competence, because other processes can produce it. A model that assigned every outcome the same score, for example, would be perfectly stable without discriminating among the frameworks at all. What counts as morally irrelevant is fixed here by human moral reasoning: a feature is irrelevant if a human reasoner would be criticised for letting it change a judgement. We treat thematic domain as morally irrelevant to framework preferences on the assumption that normative frameworks are general theories, so a principled reasoner's relative confidence in them should not depend on the area of ethical life in which a dilemma arises.

\begin{figure}[t]
  \centering
  \includegraphics[width=\textwidth]{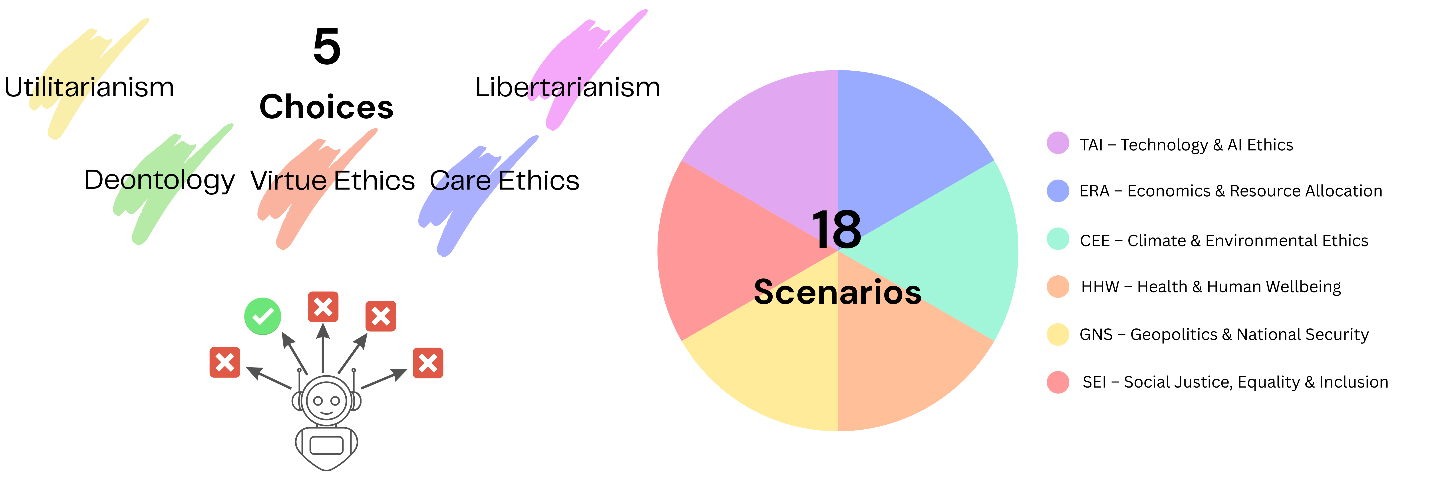}
  \caption{Overview of the evaluation dataset. The dataset consists of 18 ethical dilemma scenarios distributed evenly across six thematic domains. For each scenario, five outcomes were constructed, each recommending an action and justifying it by one of five ethical frameworks.}
  \label{fig:dataset}
\end{figure}

Three further commitments shape the design. First, we require stability across several independent dimensions of variation rather than one. A facsimile need not fail everywhere: a heuristic keyed to the wording of a prompt may be unaffected by its length. Stability along a single dimension therefore leaves such facsimiles in play. We vary three dimensions of the evaluation context (prompt length, reporting format, and thematic domain), on the assumption that conjoint stability across them is a more demanding test than stability along any one. Second, we evaluate stability at the level of framework preferences rather than verdicts. As Section~\ref{sec:related} argued, when a dilemma admits several defensible verdicts, instability in verdicts cannot distinguish a failure to track morally relevant considerations from indifference between equally defensible options, whereas a model's framework preferences can remain stable even when no single verdict is correct. Third, we adopt five canonical frameworks in normative ethics: utilitarianism, deontology, virtue ethics, care ethics, and libertarianism. We do not claim this taxonomy is exhaustive or that the frameworks are fully mutually exclusive in practice. We adopt these frameworks because they are among those most commonly used in prior LLM evaluation work, allowing comparison with existing results \citep{alizadeh2026unsupervised,chiu2025morebench,garcia2024moral,hendrycks2020aligning,keshmirian2025many,rao2023ethical,samway-etal-2025-language,thapa2025evaluation,zheng2026vispa,zhou2024rethinking}.

\subsection{Research Design}\label{sec:design}

\paragraph{Models.} We evaluate 16 configurations of 14 models from six developers; some configurations are the same model at different reasoning-effort settings: GPT-4o \citep{openai2024gpt4o}, GPT-4.1 \citep{openai2025gpt41}, o3-mini \citep{openai2025o3mini}, GPT-5 \citep[{high;}][]{singh2025openai}, GPT-5.2 \citep[{none and high;}][]{openai2025gpt52}, GPT-5.4 \citep[{none and xhigh;}][]{openai2026gpt54}, GPT-OSS-120B \citep{agarwal2025gpt}, Phi-4 \citep{abdin2024phi}, DeepSeek-V3 \citep{liu2024deepseek}, DeepSeek-R1 \citep{guo2025deepseek_r1}, DeepSeek-V3.2 \citep{liu2025deepseek}, Grok-4.1-Fast \citep[{non-reasoning;}][]{xai2025grok41}, Kimi-K2.5 \citep{team2026kimi}, and Mistral-Large-3 \citep{mistral2025mistral3}. The sample includes current frontier systems, the strongest candidates for moral competence. For GPT-5.2 and GPT-5.4, whose reasoning effort can be set to none, we use a none and a high setting (none and high for GPT-5.2, none and xhigh for GPT-5.4) and treat each setting as a separate configuration, named by the model followed by its setting, e.g., GPT-5.2 (none), since performance and behaviour can differ substantially across settings \citep{singh2025openai}. DeepSeek-V3.2 and Kimi-K2.5 were run in thinking mode, with all other settings at their defaults.

\paragraph{Dilemmas and Outcomes.} The evaluation dataset consists of 18 ethical dilemmas, three in each of six thematic domains (Figure~\ref{fig:dataset}): Economics and Resource Allocation (ERA), Climate and Environmental Ethics (CEE), Health and Human Wellbeing (HHW), Geopolitics and National Security (GNS), Social Justice, Equality, and Inclusion (SEI), and Technology and AI Ethics (TAI). These domains reflect major areas of applied ethics, including distributive justice \citep{rawls1971}, environmental and global ethics \citep{singer2011}, bioethics \citep{beauchamp2019}, just war theory \citep{walzer1977}, and information and AI ethics \citep{floridi2013}. All materials are publicly available in our accompanying repository, including the dilemma scenarios and outcomes in all three lengths, the prompts, the parsed responses and written justifications of all 16 configurations (with the raw API outputs where these were retained), and the analysis code.\footnote{\url{https://github.com/pablobiedma/Beyond-Moral-Performance}}

For every scenario, we manually constructed five alternative outcomes corresponding to the five frameworks: utilitarianism \citep{bentham1789,mill1863}, deontology \citep{kant1785}, virtue ethics \citep{aristotle2009}, care ethics \citep{gilligan1982,noddings1984}, and libertarianism \citep{locke1689,nozick1974}. Each outcome was constructed to make the framework's characteristic justification for the action it recommends explicit, while keeping the proposed action contextually plausible within the scenario.

To test stability across prompt length, each dilemma was instantiated in three progressively elaborated formats: short (15--25 words), medium (70--90 words), and long (300--340 words). The five outcomes for each dilemma were likewise written in matched short, medium, and long formats with comparable word-count ranges. The medium and long versions of each scenario and outcome were generated with a large language model from the short versions and then checked manually to ensure that they added no morally relevant content: they expand scene-setting, descriptive detail, and rhetorical texture without introducing considerations that would alter the structure of the scenario or the logic of the framework-specific justification. This yielded 54 scenario instances in total (18 dilemmas $\times$ 3 lengths), each paired with five length-matched outcomes.

Take the following short-format dilemma as an example: \emph{A city must decide whether to fund affordable housing or business incentives while facing budget constraints and rising homelessness.}

The five framework-specific outcomes are as follows:

\begin{itemize}[leftmargin=1.6em, itemsep=1pt, topsep=4pt]
  \item \textbf{Utilitarianism:} ``Fund housing to maximize societal benefit, as reducing homelessness provides greater collective welfare than business incentives that benefit fewer citizens.''
  \item \textbf{Deontology:} ``Fund housing because society has a moral duty to ensure basic needs are met before providing business incentives.''
  \item \textbf{Virtue Ethics:} ``Fund housing, demonstrating compassion and justice, virtues essential for civic leadership that prioritizes community wellbeing over economic gain.''
  \item \textbf{Care Ethics:} ``Fund housing because relationships and caring for vulnerable populations should take precedence over abstract economic benefits of businesses.''
  \item \textbf{Libertarianism:} ``Fund neither; reduce regulations on housing development and taxes on businesses, allowing market solutions with minimal government interference.''
\end{itemize}

\paragraph{Tasks.} For each scenario instance, a model configuration ranked the five outcomes from 1 (most moral) to 5 (least moral) and assigned each a morality score from 0 (least moral) to 100 (most moral). The ranking task and the scoring task were requested in two separate prompts, each presenting the dilemma with all five outcomes, so a configuration's ranks and scores come from different responses. The full prompts are available in the accompanying repository. A configuration's framework preferences are recovered from each task separately: from the scoring task, as the order of the frameworks by mean morality score, and from the ranking task, as their order by mean rank. To account for stochastic variability, each configuration was evaluated across 10 independent runs using the sampling settings specified for the respective model.\footnote{The configurations were evaluated using their respective default sampling settings; Phi-4 was the only exception, for which temperature was explicitly set to 0.7 and top-p to 0.9. In an initial collection, GPT-OSS-120B returned identical or near-identical responses across the 10 runs for almost all items; it was therefore re-collected with the same prompts and settings, and all analyses use the re-collected responses, whose runs vary as those of the other configurations do.} We did not intentionally reduce sampling stochasticity to increase reliability across runs. Given 54 scenario instances and 10 numerical outputs per instance (five ranks and five scores), this produced 540 observations per configuration per run; across 16 configurations and 10 runs, the dataset comprised 86,400 numerical data points.

\subsection{Analysis Strategy}\label{sec:analysis}

The analysis is organised around the three research questions. We use three terms in a fixed technical sense throughout. Stability is the invariance of framework preferences under variation in features that should not change a moral judgement. Agreement is the correspondence between the framework preferences recovered from the scoring and ranking tasks. Reliability is used in its psychometric sense, i.e., consistency of measurement across repeated sampling.

\paragraph{Characterising framework preferences (RQ1).} For each configuration and framework, we report descriptive statistics including mean and median morality scores, standard deviation, and the distribution of ranks. These provide the baseline characterisation of which frameworks each configuration favours.

\paragraph{Testing reliability and stability (RQ2).} The research question, whether LLMs show the reliability and stability that moral competence requires, translates into three empirical sub-questions. First, the observed framework preferences must not be artefacts of stochastic generation. We assess this separately for each reporting format. For the scoring task, Cronbach's $\alpha$ measures how consistently each outcome receives the same morality score across the 10 independent runs, treating the runs as if they were 10 raters scoring the same items; for the ranking task, Kendall's $W$ measures the consistency of the 10 rankings of the five frameworks. High values on both indicate that a configuration's framework preferences are reproducible under repeated sampling. Second, framework preferences should remain stable across prompt length and thematic domain. We assess this with two-way ANOVAs on morality scores, with prompt length and thematic domain as factors.\footnote{We checked parametric assumptions using Shapiro--Wilk tests \citep{shapiro1965analysis}, Q--Q plots, and Levene's test for variance homogeneity \citep{levene1960robust}. Residuals departed from normality for all models (Shapiro--Wilk $W = 0.86$--0.98; with 2,700 observations per model even mild departures are detected), and variances differed across thematic domains for all models and across prompt lengths for seven. We therefore repeated the tests with Kruskal--Wallis tests on framework-centred scores \citep{mckight2010kruskal}, which agree with the ANOVAs (prompt length significant in 12 of 16 models, thematic domain in all 16).} Third, the two reporting formats should yield largely the same framework preferences; if a configuration assigns the highest morality score to one framework but the best rank to another, its preferences are unstable across reporting formats. We measure agreement with Spearman's $\rho$ and Kendall's $\tau$ between the framework orders recovered from the two tasks within each configuration.\footnote{Because the scoring task and the ranking task were requested in separate prompts, agreement shows that a configuration's framework preferences survive two separate elicitations, not merely that its two reports cohere within a single response.} We report both coefficients because they capture agreement differently: Spearman's $\rho$ weights how far each framework moves between the two orders, whereas Kendall's $\tau$ counts how many of the 10 pairs of frameworks are ordered the same way (e.g., $\tau = 0.8$ means that 9 of the 10 pairs agree). For each of the three analyses, we first report the result aggregated across all 16 configurations and then break it down by configuration.

\paragraph{Comparing reasoning and non-reasoning configurations (RQ3).} To examine whether reasoning before answering is associated with greater reliability and stability, we compare configurations that produce reasoning traces before giving their answer with those that answer directly. Configurations that generate reasoning tokens before their final answer are classified as reasoning configurations, whereas those that answer directly are classified as non-reasoning configurations. The classification follows how each configuration is run rather than what its output shows, since some providers, including OpenAI, do not return the reasoning itself through the API. For models offering configurable reasoning effort, configurations with reasoning effort set to none, GPT-5.2 (none) and GPT-5.4 (none), are classified as non-reasoning, whereas configurations using explicit reasoning effort are classified as reasoning. First, we compare the central tendency of morality scores between the two groups using Mann--Whitney $U$ tests and report Cohen's $d$ as an effect-size measure. Second, we compare reliability under repeated stochastic generation using Cronbach's $\alpha$ (scoring task) and Kendall's $W$ (ranking task). Third, we compare sensitivity to prompt length and thematic domain by examining the effect sizes associated with prompt length, thematic domain, and their interaction. Finally, we compare agreement between the framework orders recovered from the scoring and ranking tasks, using Spearman's $\rho$ and Kendall's $\tau$.

%% file: sec_result.tex
\section{Results}\label{sec:results}

\subsection{Characterising Framework Preferences}\label{sec:preferences}

All 16 configurations place libertarianism last; by morality score, all but one also place care ethics and virtue ethics in the top two positions.

\begin{figure}[t]
  \centering
  \includegraphics[width=0.86\textwidth]{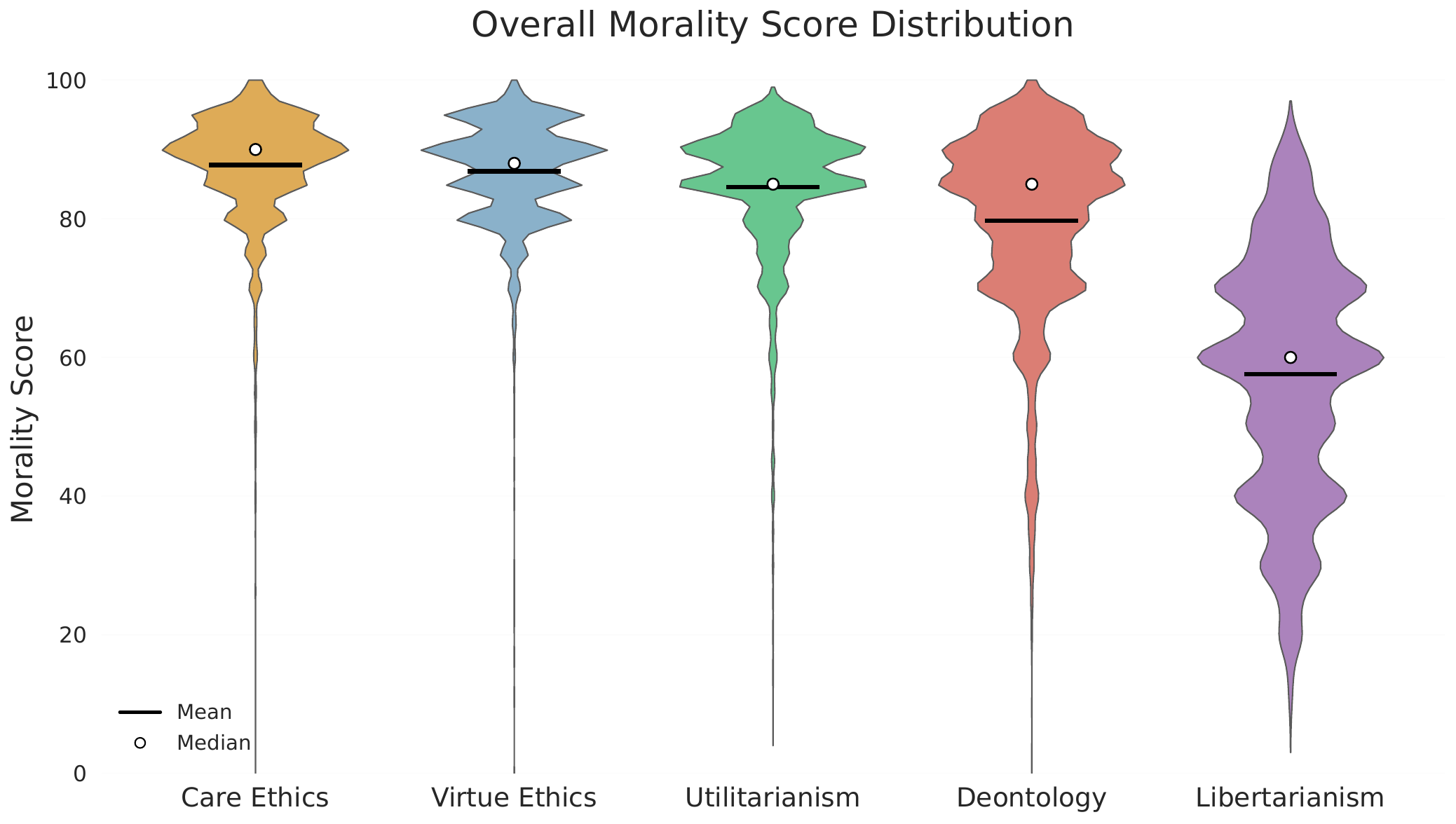}
  \caption{Distribution of morality scores across all configurations, scenarios, prompt lengths, and repeated runs for the five ethical frameworks. Care ethics receives the highest evaluations overall, followed closely by virtue ethics, utilitarianism, and deontology. Libertarianism receives substantially lower evaluations overall. Black horizontal bars indicate mean morality scores, while white circular markers indicate medians.}
  \label{fig:overall}
\end{figure}

\paragraph{Preference Ordering.} Aggregating across configurations, scenarios, and runs, the five frameworks fall into a rank order (Figure~\ref{fig:overall}). Care ethics receives the highest mean morality score (87.75), followed closely by virtue ethics (86.85). Utilitarianism (84.57) and deontology (79.76) occupy the middle positions. Libertarianism is rated substantially lower than any other framework, with a mean of 57.55, 30.20 points below care ethics.

\paragraph{Framework-Level Dispersion.} Beyond differences in central tendency, dispersion varies systematically across frameworks. Libertarian outcomes exhibit the widest spread, with scores ranging from below 30 to above 80 across scenarios and configurations. Utilitarian and deontological outcomes occupy an intermediate position, with moderate variability. Care ethics and virtue ethics display the narrowest spreads, with most scores concentrated in the upper range of the scale. The wide dispersion of libertarian scores accompanies their lower mean: libertarian outcomes are not only rated less favourably than other frameworks but also rated less consistently across scenarios.

\begin{figure}[t]
  \centering
  \includegraphics[width=\textwidth]{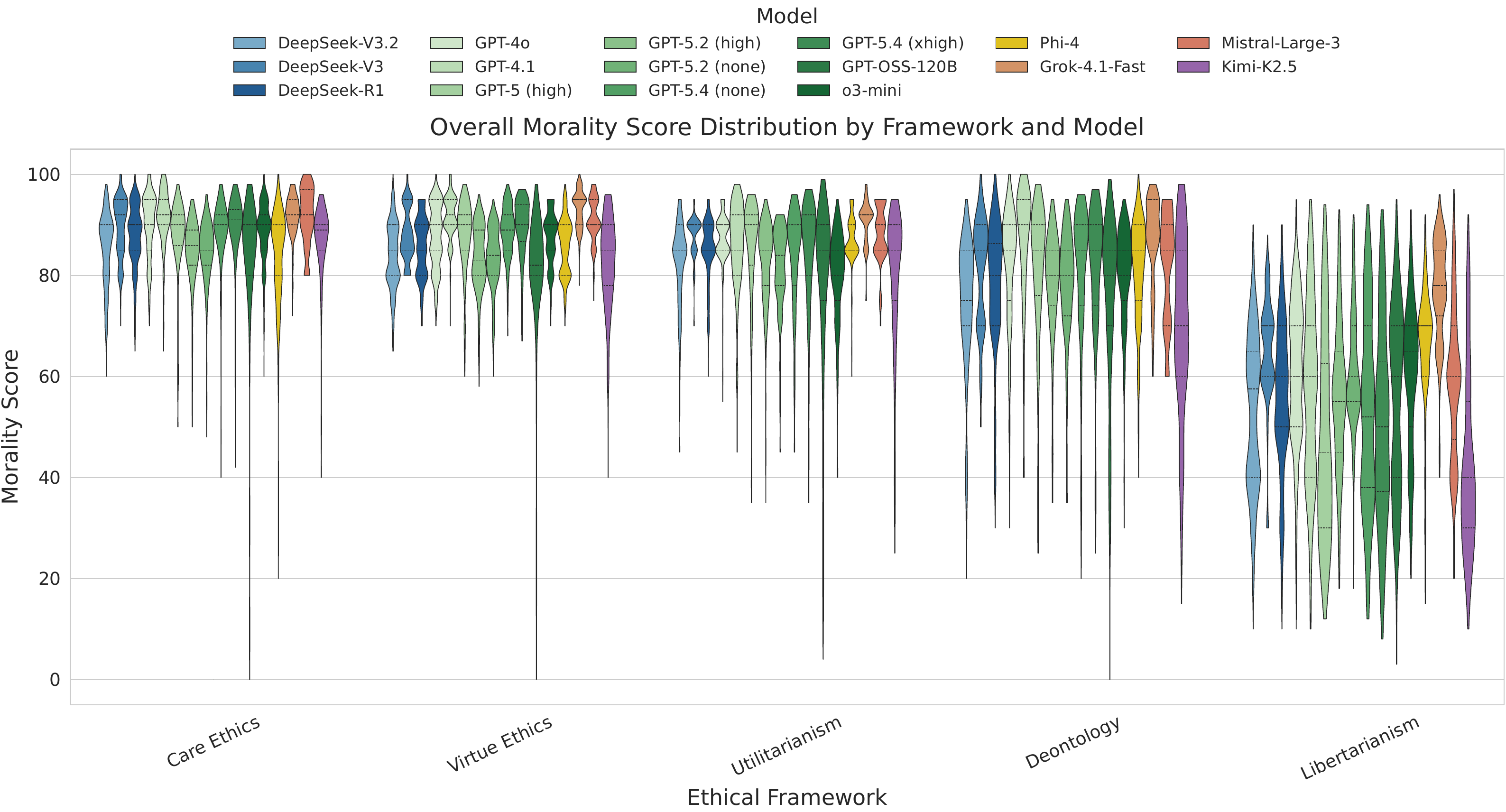}
  \caption{Each violin represents the distribution of morality scores across repeated runs for a given framework and configuration. Across configurations, care ethics and virtue ethics consistently receive the highest evaluations, while libertarian outcomes occupy a substantially lower and more variable range. Utilitarian and deontological outcomes generally occupy intermediate positions, though their relative ordering varies slightly across configurations.}
  \label{fig:by-configuration}
\end{figure}

\paragraph{Consistency across Configurations.} The preference ordering reproduces, with minor variation, in nearly every individual configuration (Figure~\ref{fig:by-configuration}). The pronounced separation between libertarian and non-libertarian outcomes appears in all 16 configurations. Care ethics and virtue ethics share the top two positions in 15 of the 16 configurations by morality score, though which of the two ranks first varies. Utilitarian and deontological outcomes occupy the two middle positions in 15 of the 16 configurations, though the gap between them varies.\footnote{More specifically, libertarian outcomes occupy the lowest-ranked position in all 16 configurations by a substantial margin, producing a clear separation between libertarian and non-libertarian frameworks. Care ethics and virtue ethics occupy the top two positions in 15 of the 16 configurations by morality score. The exception is Phi-4, where utilitarianism ranks first and care ethics ranks third. Across all 16 configurations, care ethics ranks first in 11, virtue ethics in 4, and utilitarianism in 1 (Phi-4). Utilitarianism and deontology otherwise occupy intermediate positions across models, although their relative ordering varies slightly. Specifically, utilitarianism ranks above deontology in 14 configurations, whereas deontology ranks above utilitarianism in GPT-4.1 and o3-mini. Overall, this consistency suggests that the observed preferences reflect a broadly stable pattern rather than idiosyncratic behaviour of particular configurations.}

\paragraph{Configuration-Level Variation.} Table~\ref{tab:summary} reports summary statistics for each configuration, ordered by standard deviation. Mean scores range from 72.34 (Kimi-K2.5) to 88.04 (Grok-4.1-Fast), a 16-point spread that is small relative to the 30.20-point gap between care ethics and libertarianism. Standard deviations vary more substantially across configurations, from 8.7 (Grok-4.1-Fast) to 21.1 (Kimi-K2.5). Grok-4.1-Fast is notable in combining the highest mean with the lowest standard deviation, suggesting a configuration that scores generously and discriminates relatively little between frameworks; it also rates libertarian outcomes closer to the others than any other configuration does.

\begin{table}[t]
  \centering
  \small
  \caption{Overall moral scoring tendencies and dispersion by configuration, based on morality scores. SD reflects variability across all evaluations; Range indicates the difference between the highest and lowest framework-level mean morality scores assigned by each configuration; Avg. MS denotes the mean morality score aggregated across ethical frameworks and repeated runs.}
  \label{tab:summary}
  \begin{tabular}{lrrr}
    \toprule
    Configuration & \textit{SD} & Range & Avg.\ MS \\
    \midrule
    Kimi-K2.5 & 21.06 & 43.61 & 72.34 \\
    GPT-5 (high) & 20.03 & 39.86 & 77.99 \\
    GPT-OSS-120B & 19.64 & 27.59 & 74.37 \\
    GPT-5.4 (xhigh) & 19.16 & 39.00 & 79.02 \\
    GPT-5.4 (none) & 18.31 & 36.02 & 79.00 \\
    GPT-4.1 & 16.95 & 34.55 & 82.62 \\
    DeepSeek-V3.2 & 16.17 & 32.10 & 75.73 \\
    Mistral-Large-3 & 15.56 & 33.23 & 82.30 \\
    DeepSeek-R1 & 15.44 & 32.05 & 78.62 \\
    GPT-5.2 (high) & 14.52 & 28.16 & 77.04 \\
    o3-mini & 14.05 & 28.59 & 79.23 \\
    GPT-4o & 14.04 & 28.83 & 81.70 \\
    GPT-5.2 (none) & 12.68 & 23.41 & 77.77 \\
    DeepSeek-V3 & 12.16 & 24.92 & 82.21 \\
    Phi-4 & 11.56 & 20.58 & 80.77 \\
    Grok-4.1-Fast & 8.74 & 15.82 & 88.04 \\
    \bottomrule
  \end{tabular}
\end{table}

\subsection{Reliability and Stability}\label{sec:stability}

Framework preferences are highly reliable across runs and largely preserved across reporting formats. Prompt length and thematic domain have small but statistically significant effects on raw morality scores. The effect of prompt length, however, differs between frameworks and changes the ordering among the top three.

\begin{figure}[t]
  \centering
  \includegraphics[width=0.92\textwidth]{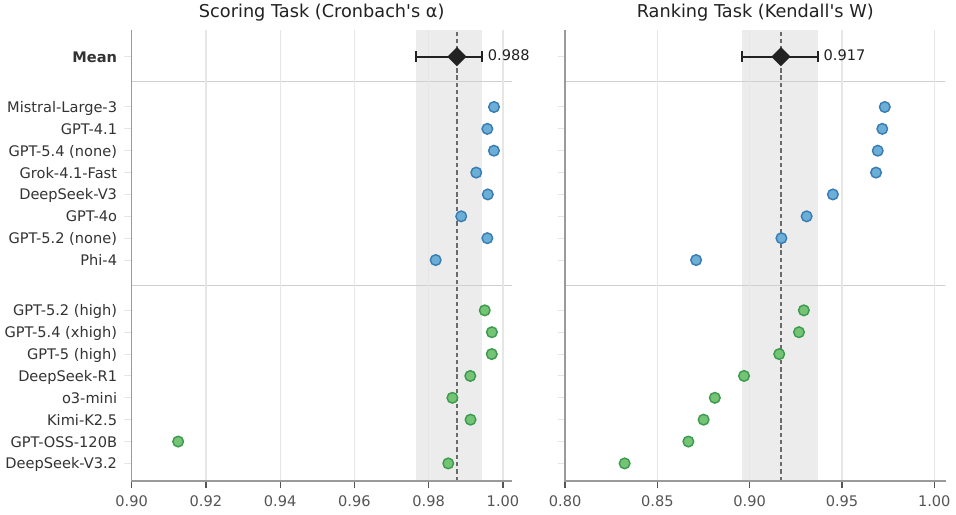}
  \caption{Reliability of framework preferences under stochastic generation. Left: Cronbach's $\alpha$ for the scoring task, measuring how consistently each outcome receives the same morality score across 10 independent runs. Right: Kendall's $W$ for the ranking task, measuring how consistently the five frameworks are ranked in the same order across the 10 runs, averaged over the 54 scenario instances. The top row shows the mean across all 16 configurations (diamond) with its 95\% bootstrap confidence interval (whisker and shaded band); dots show individual configurations, grouped into non-reasoning (blue) and reasoning (green) configurations. Cronbach's $\alpha$ ranges from 0.91 to 1.00 and Kendall's $W$ from 0.83 to 0.97, indicating highly reproducible framework preferences under repeated sampling.}
  \label{fig:reliability}
\end{figure}

\paragraph{Reliability Under Stochastic Generation.} Framework preferences are reproducible across repeated runs. Aggregated across all 16 configurations, reliability is high in both reporting formats (Figure~\ref{fig:reliability}, top row): for the scoring task, mean Cronbach's $\alpha = 0.988$ ($\mathit{SD} = 0.021$, bootstrap 95\% CI [0.976, 0.994]); for the ranking task, mean Kendall's $W = 0.917$ ($\mathit{SD} = 0.043$, 95\% CI [0.896, 0.937]). Reliability is also high for every individual configuration: Cronbach's $\alpha$ exceeds 0.97 for every configuration except GPT-OSS-120B (0.913), and Kendall's $W$ ranges from 0.83 (DeepSeek-V3.2) to 0.97 (Mistral-Large-3). Framework preferences are therefore highly consistent across repeated samples under the evaluation settings used in our experiments.

\begin{figure}[t]
  \centering
  \includegraphics[width=0.95\textwidth]{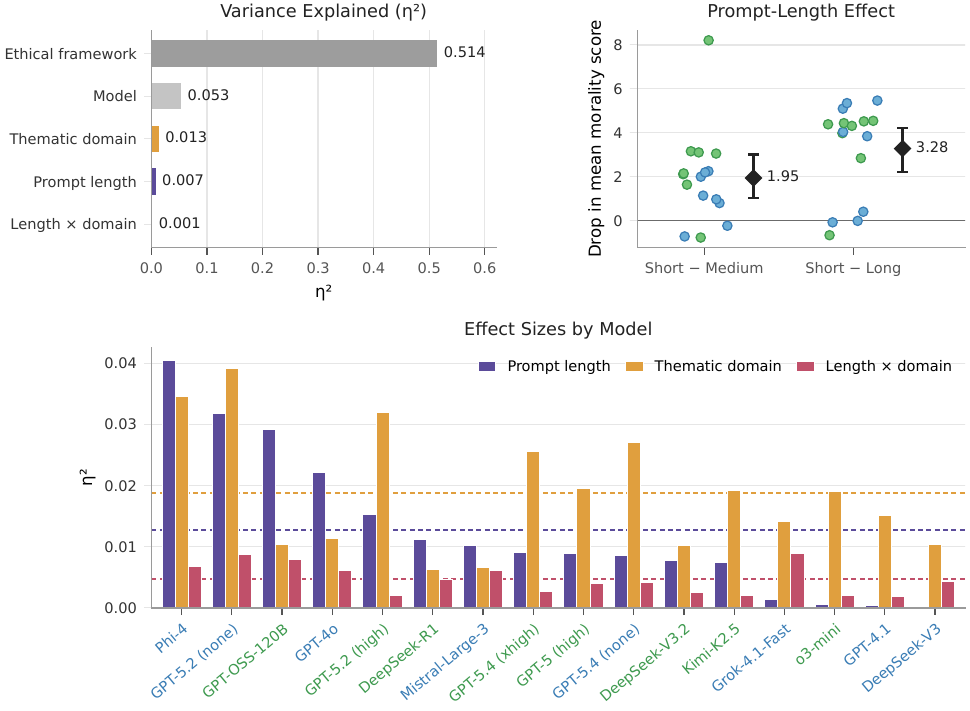}
  \caption{Effects of prompt length and thematic domain on morality scores. Top left: effect sizes ($\eta^2$) from the ANOVA aggregated across all 16 configurations; ethical framework and configuration are shown for reference. Top right: drop in each configuration's mean morality score from short to medium prompts (short $-$ medium) and from short to long prompts (short $-$ long; dots coloured by group); diamonds show the mean across configurations with 95\% bootstrap confidence intervals. Bottom: effect sizes from the per-configuration two-way ANOVAs controlling for ethical framework; dashed lines show the means across configurations (prompt length 0.013, thematic domain 0.019, interaction 0.005). Configuration names in the bottom panel are coloured by group (blue: non-reasoning; green: reasoning).}
  \label{fig:length-domain}
\end{figure}

\paragraph{Stability across Prompt Length and Thematic Domain.} Overall, morality scores shift only slightly with prompt length and thematic domain. We tested this with two-way ANOVAs including both factors and their interaction while controlling for ethical framework, first on the data aggregated across all 16 configurations and then for each configuration separately (Figure~\ref{fig:length-domain}). For the aggregate analysis, we averaged the 10 runs of each prompt condition and added configuration as a factor ($N = 4{,}320$). Together, prompt length, thematic domain, and their interaction explain only about 2\% of the variance in morality scores, compared with 51\% for ethical framework and 5\% for configuration (Figure~\ref{fig:length-domain}, top left). Prompt length ($F(2, 4283) = 38.74$, $p < .001$, $\eta^2 = 0.007$) and thematic domain ($F(5, 4283) = 27.21$, $p < .001$, $\eta^2 = 0.013$) were statistically significant, as expected with this many observations, whereas their interaction was not ($F(10, 4283) = 1.50$, $p = .133$, $\eta^2 = 0.001$). Because each thematic domain comprises different dilemmas, the domain effect also captures differences between scenarios and is best read as an upper bound on sensitivity to domain.\footnote{Kruskal--Wallis tests on framework- and configuration-centred scores lead to the same conclusion for prompt length ($H(2) = 123.8$, $p < .001$) and thematic domain ($H(5) = 108.8$, $p < .001$).} Whether the framework ordering itself is preserved across prompt lengths and thematic domains is examined at the end of this section.

Across individual configurations, effect sizes were likewise small (mean $\eta^2 = 0.013$ for prompt length, 0.019 for thematic domain, and 0.005 for their interaction; Figure~\ref{fig:length-domain}, bottom), although statistical significance varied: the effect of prompt length was significant in 13 of the 16 configurations, thematic domain in all 16, and their interaction in 11 (Table~\ref{tab:anova}, in Appendix). Interaction effects were generally smaller than the corresponding main effects.

The direction of the prompt-length effect was nevertheless highly consistent: longer prompts lowered morality scores, by about three points on the 100-point scale from short to long prompts (Figure~\ref{fig:length-domain}, top right). Aggregated across configurations, mean scores fell from 81.04 for short prompts to 79.09 for medium and 77.76 for long prompts. Taken one at a time, 13 of the 16 configurations scored lower on long than on short prompts, and 13 scored lower on medium than on short prompts (Wilcoxon signed-rank tests across configurations: $T = 7$ for long vs. short and $T = 6$ for medium vs. short, $p < .001$ in each case). Because the longer prompts are informationally redundant relative to the short versions, these findings suggest that these LLMs are sensitive to morally irrelevant presentation features, at least in the level of the scores they assign.

\begin{figure}[t]
  \centering
  \includegraphics[width=0.95\textwidth]{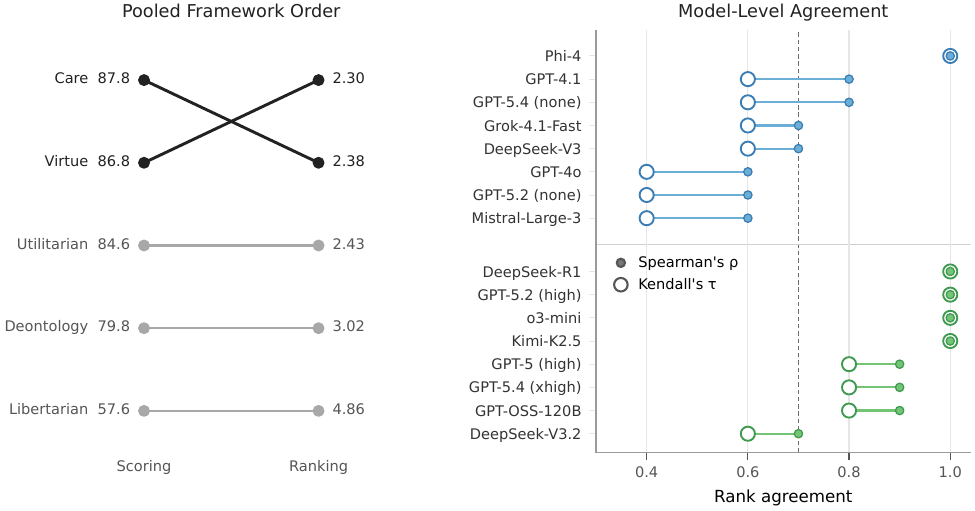}
  \caption{Agreement between the framework orders obtained directly from the ranking task and those recovered from morality scores. Left: aggregated across all 16 configurations, frameworks ordered by mean morality score (left, with mean score) and by mean direct rank (right, with mean rank); only care and virtue ethics change positions ($\rho = 0.90$, $\tau = 0.80$). Right: configuration-level Spearman's $\rho$ (filled) and Kendall's $\tau$ (open) for non-reasoning (blue) and reasoning (green) configurations; concentric markers indicate $\rho = \tau = 1.0$, and the dashed line marks a common threshold for strong agreement ($\rho = 0.7$).}
  \label{fig:format-agreement}
\end{figure}

\paragraph{Agreement between Reporting Formats.} The two reporting formats yield nearly the same framework preferences. We compared the preferences obtained directly from the ranking task with those recovered from morality scores. Aggregated across all 16 configurations, the two formats produce the same order except that care ethics and virtue ethics swap the top two positions (Figure~\ref{fig:format-agreement}, left), yielding $\rho = 0.90$ and $\tau = 0.80$.

At the configuration level, most configurations show strong agreement between the two formats (Figure~\ref{fig:format-agreement}, right; Table~\ref{tab:agreement}). Thirteen of the 16 configurations achieve $\rho \geq 0.7$, a common threshold for strong agreement \citep{schober2018correlation}, eight achieve $\tau \geq 0.7$, and five show perfect agreement ($\rho = \tau = 1.0$). By contrast, GPT-4o, GPT-5.2 (none), and Mistral-Large-3, all non-reasoning configurations, show the lowest agreement ($\rho = 0.6$, $\tau = 0.4$). However, Phi-4, a non-reasoning configuration, achieves perfect agreement ($\rho = \tau = 1.0$), a result replicated in an independent re-run, which shows that reasoning before answering is not necessary for high agreement. Importantly, even in configurations with lower agreement, the broad preference ordering identified in Section~\ref{sec:preferences} remains largely intact: libertarian outcomes consistently occupy the lowest-ranked position, while care ethics, virtue ethics, and utilitarianism share the upper range. The differences therefore concern finer-grained ordering among the non-libertarian frameworks rather than a complete change of the overall preference ordering.

\paragraph{Framework-Specific Effects of Prompt Length and Domain.} The analyses above ask how much prompt length and thematic domain shift morality scores overall. Here we ask whether they shift the scores of some ethical frameworks more than others, and therefore whether the framework ordering is preserved across prompt lengths and domains. For this analysis, morality scores were averaged over the 10 runs for each configuration, dilemma, prompt length, and framework ($16 \times 18 \times 3 \times 5 = 4{,}320$ scores). The effect of prompt length differs markedly between frameworks. The framework $\times$ length interaction is significant in repeated-measures ANOVAs with configurations ($F(8, 120) = 34.00$, $\eta_p^2 = 0.69$) or dilemmas ($F(8, 136) = 9.78$, $\eta_p^2 = 0.37$) as the unit of analysis (Greenhouse--Geisser corrected $p < .001$ in both cases), and when configurations and dilemmas are treated as random factors at the same time \citep[{min $F'(8, 206) = 7.59$, $p < .001$;}][]{clark1973language}. From short to long prompts, mean scores fell by 10.02 points for libertarian outcomes, 5.04 for virtue ethics outcomes, and 3.38 for care ethics outcomes; each decrease appeared in at least 15 of the 16 configurations and 16 of the 18 dilemmas (Wilcoxon signed-rank tests, Holm-corrected across frameworks, $p < .01$ in each case). Scores for utilitarian (+1.37) and deontological (+0.66) outcomes changed little, and neither change was consistent across dilemmas. Medium prompts fall in between (libertarianism $-6.59$, virtue ethics $-3.64$, care ethics $-1.27$).

As a result, the framework ordering is only partly preserved across prompt lengths. With short prompts, care ethics and virtue ethics occupy the top two positions in all 16 configurations; with medium prompts they do so in 10 and with long prompts in 7. With long prompts, care ethics (85.92), utilitarianism (85.27), and virtue ethics (84.70) no longer differ significantly from one another (Wilcoxon signed-rank tests across configurations, Holm-corrected $p \geq .06$), although all three remain above deontology (79.82) and libertarianism (53.07). Libertarian outcomes receive the lowest scores in every configuration at every prompt length. Within configurations, the framework orders recovered from short and long prompts agree less (mean Kendall's $\tau = 0.65$) than the orders recovered from two halves of the runs at the same prompt length ($\tau = 0.97$), so the change in order is not run-to-run noise. Because the medium and long versions of each dilemma also elaborate the descriptions of the five outcomes (Section~\ref{sec:design}), these framework-specific changes may partly reflect how the elaborated outcomes are worded rather than prompt length as such.

Thematic domain, by contrast, has no effect on framework scores that can be separated from differences between dilemmas. Because each domain comprises different dilemmas, a domain effect must be judged against the differences between dilemmas within the same domain. With the 18 dilemmas as the unit of analysis and thematic domain as a between-dilemma factor (scores averaged across configurations, prompt lengths, and runs), neither thematic domain ($F(5, 12) = 1.04$, $p = .44$) nor its interaction with ethical framework ($F(20, 48) = 0.50$, $p = .90$, Greenhouse--Geisser corrected) is significant, and libertarian outcomes receive the lowest scores in all 96 combinations of configuration and domain. With only three dilemmas per domain this test has little power, so a modest effect of domain cannot be ruled out, but we find no evidence that the framework ordering depends on domain.

\subsection{Reasoning and Non-Reasoning Configurations}\label{sec:reasoning}

We compare reasoning and non-reasoning configurations (Table~\ref{tab:classification}) on four measures: (i) overall morality scores, (ii) reliability across repeated runs, (iii) sensitivity to prompt length and thematic domain, and (iv) agreement between the two reporting formats. Reasoning configurations agree more closely across formats but rank less reliably across runs; on the other measures of reliability and stability, we detect no difference.

\begin{table}[t]
  \centering
  \small
  \caption{Classification of configurations into non-reasoning and reasoning groups.}
  \label{tab:classification}
  \begin{tabular}{ll}
    \toprule
    Non-Reasoning Configurations & Reasoning Configurations \\
    \midrule
    GPT-4o & o3-mini \\
    GPT-4.1 & GPT-5 (high) \\
    Phi-4 & GPT-5.2 (high) \\
    DeepSeek-V3 & GPT-5.4 (xhigh) \\
    Grok-4.1-Fast & DeepSeek-R1 \\
    Mistral-Large-3 & DeepSeek-V3.2 \\
    GPT-5.2 (none) & GPT-OSS-120B \\
    GPT-5.4 (none) & Kimi-K2.5 \\
    \bottomrule
  \end{tabular}
\end{table}

\begin{figure}[t]
  \centering
  \includegraphics[width=\textwidth]{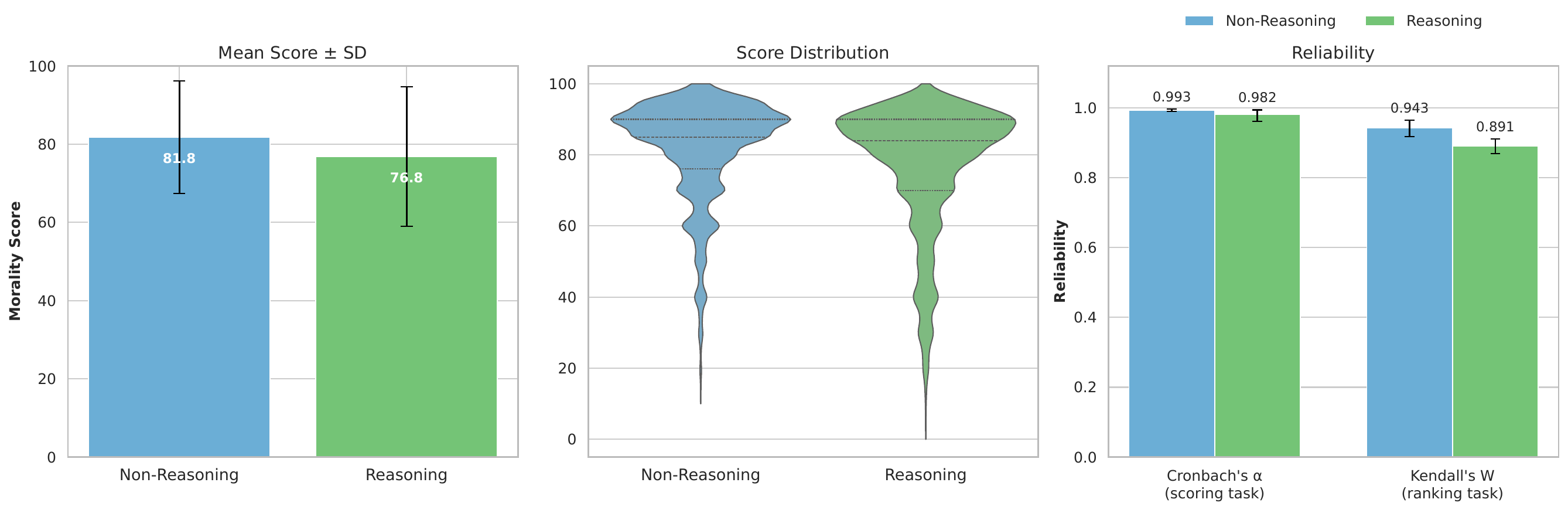}
  \caption{Comparison of non-reasoning and reasoning configurations across three complementary measures of moral evaluation, aggregated over all ethical dilemmas, prompt lengths, and repeated runs. Left: Mean morality score. Non-reasoning configurations assign higher morality scores overall than reasoning configurations, while preserving the same broad ordering of ethical frameworks. Centre: Distribution of morality scores across all evaluations. Right: consistency when the same prompt is asked 10 times, measured by Cronbach's $\alpha$ for the scoring task and Kendall's $W$ for the ranking task (means across the configurations in each group, with 95\% bootstrap confidence intervals); both groups are highly consistent, although reasoning configurations, compared to the non-reasoning ones, are significantly less consistent on the ranking task.}
  \label{fig:reasoning}
\end{figure}

First, non-reasoning configurations produce higher overall morality scores ($M = 81.80$) than reasoning configurations ($M = 76.79$; Figure~\ref{fig:reasoning}, left). This difference is statistically significant (Mann--Whitney $U = 6.0$, $p = .005$), and the effect is large (Cohen's $d = -1.81$). It concerns the level of the scores rather than their stability, and both groups retain the same broad preference ordering identified in Section~\ref{sec:preferences}.

Second, we compared reliability across repeated runs (Figure~\ref{fig:reasoning}, right). For the scoring task, Cronbach's $\alpha$ is very high in both groups (0.993 for non-reasoning and 0.982 for reasoning configurations), and the difference is not statistically significant (Mann--Whitney $U = 21.0$, $p = .279$, Cohen's $d = -0.55$). For the ranking task, Kendall's $W$ is high in both groups but lower for reasoning configurations (0.891) than for non-reasoning configurations (0.943), and this difference is statistically significant ($U = 8.0$, $p = .010$, $d = -1.52$). Reasoning configurations therefore reproduce their framework rankings somewhat less consistently across repeated runs; the two lowest values of $W$ in the sample both belong to reasoning configurations (DeepSeek-V3.2, 0.83; GPT-OSS-120B, 0.87; Figure~\ref{fig:reliability}).

Third, we compared the ANOVA effect sizes for prompt length, thematic domain, and their interaction between the two groups. For prompt length, the mean $\eta^2$ was 0.0144 for non-reasoning configurations and 0.0112 for reasoning configurations (Mann--Whitney $U = 32.0$, $p > .99$; Cohen's $d = -0.26$). For thematic domain, the corresponding means were 0.0198 and 0.0178 ($U = 29.0$, $p = .798$; $d = -0.19$). For the prompt-length $\times$ thematic-domain interaction, mean $\eta^2$ was 0.0059 for non-reasoning configurations and 0.0035 for reasoning configurations ($U = 15.0$, $p = .083$; $d = -1.07$). None of the three differences is statistically significant, and the prompt-length effect points in the same direction in both groups: long prompts lowered scores by 3.02 points on average for non-reasoning configurations and by 3.55 points for reasoning configurations ($U = 32.0$, $p > .99$; Figure~\ref{fig:length-domain}, top right). We therefore detect no difference between the groups in their sensitivity to prompt length and thematic domain.

\begin{figure}[t]
  \centering
  \includegraphics[width=0.5\textwidth]{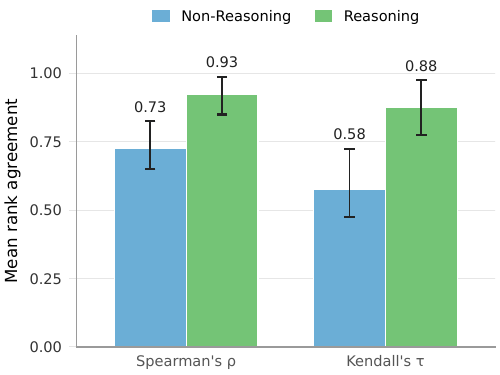}
  \caption{Agreement between the framework orders recovered from the ranking and scoring tasks for non-reasoning (blue) and reasoning (green) configurations. Bars show the mean Spearman's $\rho$ and Kendall's $\tau$ across the configurations in each group; error bars show 95\% bootstrap confidence intervals across configurations.}
  \label{fig:format-agreement-groups}
\end{figure}

Finally, we compared agreement between the two reporting formats. Mean Spearman agreement was substantially higher among reasoning configurations than among non-reasoning configurations (0.93 vs. 0.73; Figure~\ref{fig:format-agreement-groups}), and this difference was statistically significant (Mann--Whitney $U = 55.0$, $p = .015$; Cohen's $d = 1.63$). Kendall's $\tau$ yielded the same pattern, with mean agreement of 0.88 for reasoning configurations and 0.58 for non-reasoning configurations ($U = 56.0$, $p = .010$). Reasoning configurations therefore show closer agreement between the framework orders recovered from the scoring and ranking tasks.

In addition, we asked whether the framework-specific effect of prompt length described in Section~\ref{sec:stability} differs between reasoning and non-reasoning configurations. We find no evidence that it does. Both groups show the same pattern: from short to long prompts, libertarian outcomes lost 11.65 points among reasoning configurations and 8.40 points among non-reasoning configurations, virtue ethics outcomes 5.45 and 4.63 points, and care ethics outcomes 2.85 and 3.91 points, whereas utilitarian and deontological scores changed by less than two points in either group. With configuration as the unit of analysis, neither the group $\times$ framework $\times$ length interaction ($F(8, 112) = 1.27$, $p = .27$) nor the group $\times$ length interaction ($F(2, 28) = 1.67$, $p = .17$) is significant, and neither is the group $\times$ framework $\times$ domain interaction ($F(20, 280) = 1.03$, $p = .35$); $p$-values come from exact permutation tests over all 12,870 ways of dividing the 16 configurations into two groups of eight. No framework's change differs between the groups after Holm correction; the largest difference concerns libertarianism, whose scores fell 3.25 points more among reasoning configurations (Mann--Whitney $U = 16.5$, $p = .130$ uncorrected; Cohen's $d = -0.74$). Care ethics and virtue ethics occupy the top two positions with short prompts in all eight configurations of each group, but with long prompts in only four reasoning and three non-reasoning configurations, and libertarian outcomes receive the lowest scores in every configuration of both groups at every prompt length.

Two models allow a cleaner comparison, because they were run with reasoning both switched off and switched on: GPT-5.2 (none vs. high) and GPT-5.4 (none vs. xhigh). For each framework, we compared the short-to-long change between the two settings across the 18 dilemmas (Wilcoxon signed-rank tests, Holm-corrected across frameworks). For GPT-5.2, switching reasoning on did not significantly change the length effect for any framework. For GPT-5.4, it did not significantly change the length effect for care ethics, virtue ethics, utilitarianism, or deontology, but it enlarged the drop for libertarian outcomes from 9.56 to 15.69 points (larger in 15 of 18 dilemmas, $p = .012$). Because the two models disagree, we do not attribute this difference to reasoning as such. With eight configurations per group, these tests have limited power: they show no evidence of a difference between reasoning and non-reasoning configurations, rather than evidence that none exists.

%% file: sec_discuss.tex
\section{Discussion}\label{sec:discussion}

This section discusses what our findings establish. Section~\ref{sec:regularity} characterizes the moral orientation shared across the sixteen configurations as a descriptive regularity and draws two implications: a moral monoculture across deployed frontier systems, and the opaque and nondemocratic character of the process that installs it. Section~\ref{sec:competence} argues that stability results license a directional update toward the presence of moral competence. Section~\ref{sec:reasoning-discussion} takes up the difference between reasoning and non-reasoning models. Section~\ref{sec:limitations} sets out the limitations of the study.

\subsection{A Robust Descriptive Regularity}\label{sec:regularity}

Aggregated across models, scenarios, prompt lengths, and runs, the five frameworks fall into one order: care and virtue ethics at the top, utilitarianism and deontology in the middle, libertarianism far below the rest. The gap between the top and bottom frameworks is 30.20 points, roughly twice the spread between the most and least generous models. It is not an artifact of aggregation: libertarian outcomes rank last in all sixteen configurations, and care and virtue occupy the top two positions in fifteen of sixteen by morality score; in the direct ranking task (Table~\ref{tab:ranking}, in Appendix) they do so in seven, and in the remaining nine utilitarianism enters the top two. The order is reproducible under repeated sampling and largely preserved when the reporting format changes from scoring to ranking, and we find no evidence that it depends on thematic domain. It is not invariant across prompt length, however: libertarian outcomes rank last at every length, but with long prompts care ethics, utilitarianism, and virtue ethics are no longer distinguishable (Section~\ref{sec:stability}).

What produces the shared orientation cannot be answered by our design. Two general and non-exclusive sources are possible. (1) The pretraining corpus may already carry the pattern: human moral writing may appeal to care and character more often than to libertarian non-interference. Machine learning may also amplify the pattern: training is by default optimized for majority patterns, so less common patterns are learned poorly, and the output distribution ends up more homogeneous than the distribution it was trained on \citep{huang2025generative}. (2) Reinforcement learning with human feedback (RLHF) may embed this orientation. RLHF optimizes the model toward outputs that human raters score highly. The helpful, honest, and harmless principle that guides RLHF \citep{askell2021general,bai2022training} is not neutral among our five frameworks: harmlessness and helpfulness are associated with care ethics, and honesty and compassion are standardly listed among the virtues. To the extent that raters apply this principle, the moral orientation we observe can be built into the models.

One model-level pattern is worth highlighting. Grok-4.1-Fast compresses the gap between libertarianism and the other four frameworks more than any other configuration in our sample, which is consistent with xAI's stated positioning around reduced left-leaning bias in its training pipeline \citep{xai2025grok4fast}. That said, it also returns the highest mean score (88.04) and the smallest framework-level range (15.82) in the sample, so it discriminates less among all five frameworks and not only at the libertarian end.

The preference for care and virtue is a descriptive regularity, not a prescriptive one. Whether convergence on this hierarchy is good turns on normative questions we do not take up here. What does follow from the regularity is a pair of concerns about how the orientation is distributed across deployed systems, and about who chose it.

\paragraph{Moral monoculture across frontier systems.} Convergence of moral orientation means that the diversity available across deployed systems is low. The familiar worry about embedded values is that the orientation might be wrong. The worry here is different: whatever the shared hierarchy's blind spots are, they are systemic, not model-specific. A user who objects to how one system handles value judgment has nowhere better to go.

\paragraph{The non-democratic process that embedded it.} Our data identify the orientation without localizing its source because the information needed is not public: neither corpus composition, nor annotator instructions, nor constitutional principles are completely disclosed for the models we tested. A small group within each company gets to decide which values the system embeds, and that decision is not open to challenge by its users. Essentially, it is an authoritarian rather than a democratic process for value alignment \citep{huang2025democratizing}.

\subsection{The Data Are More Compatible with Moral Competence than with Mere Performance}\label{sec:competence}

As discussed in Section~\ref{sec:intro}, there is an asymmetry in our interpretation framework. Instability across multiple variations in morally irrelevant features is strong evidence against the presence of moral competence; in contrast, stability across those same variations is only weak evidence for competence. This is because stability has many possible producers, only some of which involve competence. Accordingly, our findings should be understood as providing asymmetric evidence: instability provides stronger grounds for lowering credence in moral competence, whereas stability provides only limited grounds for increasing it.

Across the three stability assessments, the models show high reliability under repeated sampling and limited but non-zero sensitivity to morally irrelevant perturbations. (1) Reliability under stochastic generation is high for every configuration ($\alpha \geq 0.91$, $W \geq 0.83$). (2) The overall effects of prompt length and thematic domain on morality scores were small (together about 2\% of the variance), although they were statistically significant in most models; the effect of prompt length, however, fell mainly on libertarian, virtue ethics, and care ethics outcomes and changed the ordering among the top three frameworks (Section~\ref{sec:stability}). (3) Framework preferences elicited by ranking agree with those reconstructed from scores in thirteen of sixteen models. Overall, the sensitivity to these features appears limited in magnitude, although it is not uniformly absent across models and, in the case of prompt length, reaches the ordering of the top frameworks. Read through the methodological framework adopted, this pattern is, on balance, more compatible with the presence of underlying moral competence than with its absence; the dependence of the ordering on prompt length is the main evidence on the other side.

Sensitivity of this degree does not undermine the attribution of competence, because the default reference system, human cognition, does not avoid it either. For example, human moral judgment is sensitive to the order in which cases are presented and to how outcomes are described (i.e., as lives saved vs. lives lost); moreover, such sensitivity is not eliminated by philosophical expertise. \citet{schwitzgebel2012expertise} found that professional philosophers' judgments about double-effect, action--omission, and moral luck cases shifted with presentation order by about as much as non-philosophers' did. In a follow-up study, neither an enforced delay for reflection nor self-reported expertise reduced the effects \citep{schwitzgebel2015philosophers}. Presentation order and outcome framing are as morally irrelevant as the features we manipulate here. Perfect stability therefore cannot be the criterion for moral competence, since adopting it would disqualify moral competence in humans.

Where the sensitivity comes from is a question we cannot answer. (1) The model may itself track morally irrelevant features, and the sensitivity marks a limit on its moral competence. (2) The sensitivity may be introduced downstream of the model. Deployed LLMs sit inside production stacks that include components we cannot observe, such as content filters and other serving-side components, which could respond to prompt length or subject matter \citep{dong2024position,xai2025grok41}. Providers may also modify the served model without notice \citep{gao2025model}. Since we query the stack, not the model alone, our behavioral data cannot distinguish (1) from (2).

One qualification: we have read the results as evidence for a human-like moral competence with some performance error. Yet, a different interpretation is compatible with the data: a non-human-like moral competence, functioning properly, on features we have classified as irrelevant. Our design cannot discriminate between the two interpretations, as the designation ``morally irrelevant'' is defined over what human moral reasoning would be criticized for tracking. For most purposes in AI ethics, the concern is whether these systems reason morally in a human-like way. Whether alien moral competence might legitimately be sensitive to features we find morally irrelevant is a question we cannot address here.

Two implications follow. First, whatever its source, the sensitivity is a defect in deployment. The direction of the length effect is consistent across systems: longer prompts lower morality scores. A system that judges a case more harshly because the case was described at greater length can lead to problematic moral decision-making in deployment. This issue is made worse by the trend of increasing prompt length, with frontier context windows expanding roughly thirtyfold per year since 2023 \citep{burnham2025llms}. Other such features presumably exist. Second, if the non-human-like competence interpretation turns out to be correct, the problem worsens. A human-like moral competence with some sensitivity means that its moral judgments remain predictable, even if they depart from our predictions to a small degree. A system with non-human-like moral competence, on the other hand, would produce moral judgments that depart systematically, in ways that are hard to predict.

\subsection{Reasoning Affects Some Dimensions of Judgment Stability but Not Others}\label{sec:reasoning-discussion}

Model configurations that emit an explicit reasoning trace, compared to non-reasoning configurations, show greater stability across reporting formats but lower reliability when performing the ranking task. On the remaining measures we detect no difference: reliability on the scoring task is high in both groups and does not differ significantly between them, and the effect sizes for prompt length, thematic domain, and their interaction do not differ significantly. Reasoning models do assign lower morality scores overall (Section~\ref{sec:reasoning}), but this concerns the raw scores rather than their stability. The two significant differences in consistency point in opposite directions: reasoning before answering brings a model's direct ranking closer to the order implied by its own scores, yet the ranking itself varies more from run to run (Kendall's $W$ 0.891 vs. 0.943).

One relevant finding to note is that reasoning also does not always improve stability in human subjects. \citet{schwitzgebel2015philosophers}, discussed in Section~\ref{sec:competence}, imposed a fifteen-second delay and instructed participants to reflect before answering. Framing and order effects were not smaller in that condition.

The result bears on an assumption now common in AI research: that reasoning architectures should improve performance on difficult tasks. They do perform better on mathematical and scientific benchmarks \citep{jaech2024openai,guo2025deepseek_r1}, where answers are scored against ground truth and the trace can be trained toward a verifiable correct answer. Our probe asks whether a system agrees with itself, not whether it is right. What those benchmarks measure and what we measure may therefore come apart, and the absence of a consistent gain here may say more about the limit of our instrument than about moral reasoning. Suppose, though, that the absence is real. Two further explanations are worth pursuing, and they are not exclusive. Something about moral reasoning, or about practical reasoning generally, may resist improvement by an explicit reasoning process. Or the training data for moral reasoning may be too thin at present to support a model that reasons its way to greater consistency.

\subsection{Limitations}\label{sec:limitations}

Our analysis rests on a competence/performance distinction borrowed from linguistics. The distinction is contested in its home discipline, and extending it to moral cognition, let alone to non-human systems, is a substantive theoretical commitment.

A further set of limitations concerns what we measured. Each result is a fact about a deployed system at a moment in time, since a query reaches a model together with a system prompt, content filters, and routing, none of which is disclosed or stable. Providers revise these components without notice and deprecate configurations, so the study is not reproducible in the ordinary sense and the unit of analysis is not strictly the model. The sixteen configurations are also less independent of one another than their number suggests. All are frontier systems from a small number of labs, they share training corpora and architectural lineage, and several are the same base model at different reasoning-effort settings. Convergence across systems related in these ways is weaker evidence than convergence across genuinely independent ones.

The design constrains the comparisons we can draw. Eight configurations against eight leave Section~\ref{sec:reasoning-discussion} able to detect only large effects, so its null results are failures to detect rather than strong evidence of equivalence. The short, medium, and long versions of each dilemma were also LLM-generated (and manually checked), and if the longer versions nevertheless introduced morally relevant content despite the elaborative design, the length effect may not be a sensitivity to an irrelevant feature. This applies in particular to the framework-specific length effects reported in Section~\ref{sec:stability}, because the outcome descriptions were elaborated together with the scenarios.

The probe itself is narrow in three respects. Eighteen dilemmas across six domains exhaust neither applied ethics nor the range of situations in which these systems will be asked to judge. The five frameworks are a reference set inherited from prior work rather than a taxonomy we defend. Finally, scores and rankings over predefined outcomes capture a thin slice of how a moral orientation could manifest.

%% file: sec_conclu.tex
\section{Conclusion}\label{sec:conclusion}

We asked whether the moral framework preferences of deployed LLMs survive variation in features that should not change a moral judgment. Across sixteen frontier configurations they largely survive. The same hierarchy appears in nearly every configuration we tested: care and virtue ethics above utilitarianism and deontology (in 15 of 16 configurations), and all four far above libertarianism (in all 16); it holds under repeated sampling and across thematic domains, and largely across two reporting formats. Across contextual elaboration it holds only in part: libertarianism remains last at every prompt length, but with long prompts care ethics, utilitarianism, and virtue ethics can no longer be distinguished. Reasoning before answering is associated with closer agreement between those two formats, although reasoning configurations reproduce their rankings less consistently across repeated runs. On balance, this pattern favors the presence of moral competence over its absence, with the sensitivity to prompt length as the main qualification. At the same time, the hierarchy these systems share was installed by training processes that no one outside the developing labs can inspect or contest.

%% file: sec_statements.tex
\section*{Disclosure Statement}

The authors report there are no competing interests to declare.

\section*{Funding}

This research received no specific grant from any funding agency in the public, commercial, or not-for-profit sectors.

\section*{Data Availability Statement}

The dilemma scenarios, prompts, parsed model responses (with raw outputs where retained), and analysis code supporting the findings of this study are openly available at \url{https://github.com/pablobiedma/Beyond-Moral-Performance}.

%% file: appendix.tex
\appendix

\section*{Appendix}\label{sec:appendix}

\begin{table}[ht]
  \centering
  \small
  \caption{Two-way ANOVA results (Prompt Length $\times$ Thematic Domain) for each model, controlling for ethical framework (six thematic domains; ten runs per prompt condition). Reported values include $p$-values and effect sizes ($\eta^2$) for prompt length, thematic domain, and their interaction. The ANOVA formulation was: $\text{score} \sim C(\text{Version}) + C(\text{Theme}) + C(\text{Framework}) + C(\text{Version}){:}C(\text{Theme})$. Because the ten runs of each prompt condition enter as separate observations, these per-model $p$-values are best read as descriptive; the aggregate analysis in Section~\ref{sec:stability} averages runs before testing.}
  \label{tab:anova}
  \begin{tabular}{lrrrrrr}
    \toprule
    Model & $p_{\text{Length}}$ & $p_{\text{Theme}}$ & $p_{\text{Interaction}}$ & $\eta^2_{\text{Length}}$ & $\eta^2_{\text{Theme}}$ & $\eta^2_{\text{Interaction}}$ \\
    \midrule
    Grok-4.1-Fast & 0.027 & $<10^{-4}$ & $<10^{-4}$ & 0.001 & 0.014 & 0.009 \\
    GPT-5.2 (none) & $<10^{-4}$ & $<10^{-4}$ & $<10^{-4}$ & 0.032 & 0.039 & 0.009 \\
    GPT-OSS-120B & $<10^{-4}$ & $<10^{-4}$ & 0.0008 & 0.029 & 0.010 & 0.008 \\
    Phi-4 & $<10^{-4}$ & $<10^{-4}$ & 0.0001 & 0.041 & 0.035 & 0.007 \\
    GPT-4o & $<10^{-4}$ & $<10^{-4}$ & $<10^{-4}$ & 0.022 & 0.011 & 0.006 \\
    Mistral-Large-3 & $<10^{-4}$ & $<10^{-4}$ & $<10^{-4}$ & 0.010 & 0.007 & 0.006 \\
    DeepSeek-R1 & $<10^{-4}$ & $<10^{-4}$ & 0.0003 & 0.011 & 0.006 & 0.005 \\
    DeepSeek-V3 & 0.820 & $<10^{-4}$ & 0.002 & 0.000 & 0.010 & 0.004 \\
    GPT-5.4 (none) & $<10^{-4}$ & $<10^{-4}$ & 0.003 & 0.009 & 0.027 & 0.004 \\
    GPT-5 (high) & $<10^{-4}$ & $<10^{-4}$ & 0.002 & 0.009 & 0.019 & 0.004 \\
    GPT-5.4 (xhigh) & $<10^{-4}$ & $<10^{-4}$ & 0.037 & 0.009 & 0.026 & 0.003 \\
    DeepSeek-V3.2 & $<10^{-4}$ & $<10^{-4}$ & 0.091 & 0.008 & 0.010 & 0.003 \\
    Kimi-K2.5 & $<10^{-4}$ & $<10^{-4}$ & 0.190 & 0.007 & 0.019 & 0.002 \\
    GPT-5.2 (high) & $<10^{-4}$ & $<10^{-4}$ & 0.212 & 0.015 & 0.032 & 0.002 \\
    o3-mini & 0.180 & $<10^{-4}$ & 0.272 & 0.001 & 0.019 & 0.002 \\
    GPT-4.1 & 0.280 & $<10^{-4}$ & 0.228 & 0.000 & 0.015 & 0.002 \\
    \bottomrule
  \end{tabular}
\end{table}

\begin{table}[ht]
  \centering
  \footnotesize
  \caption{Framework preference rankings derived from direct ranking tasks. Frameworks are listed from 1st (most preferred) to 5th by their mean rank, averaged across dilemmas, prompt lengths, thematic domains, and repeated runs.}
  \label{tab:ranking}
  \begin{tabular}{lccccc}
    \toprule
    Model & 1st & 2nd & 3rd & 4th & 5th \\
    \midrule
    DeepSeek-R1 & Care Ethics & Virtue Ethics & Utilitarianism & Deontology & Libertarianism \\
    DeepSeek-V3.2 & Virtue Ethics & Utilitarianism & Care Ethics & Deontology & Libertarianism \\
    GPT-4.1 & Virtue Ethics & Care Ethics & Utilitarianism & Deontology & Libertarianism \\
    GPT-4o & Utilitarianism & Virtue Ethics & Care Ethics & Deontology & Libertarianism \\
    GPT-5.2 (high) & Care Ethics & Virtue Ethics & Utilitarianism & Deontology & Libertarianism \\
    GPT-5.2 (none) & Utilitarianism & Virtue Ethics & Care Ethics & Deontology & Libertarianism \\
    GPT-5.4 (none) & Care Ethics & Virtue Ethics & Deontology & Utilitarianism & Libertarianism \\
    GPT-5.4 (xhigh) & Care Ethics & Virtue Ethics & Utilitarianism & Deontology & Libertarianism \\
    GPT-5 (high) & Virtue Ethics & Utilitarianism & Care Ethics & Deontology & Libertarianism \\
    Grok-4.1-Fast & Utilitarianism & Virtue Ethics & Care Ethics & Deontology & Libertarianism \\
    Kimi-K2.5 & Care Ethics & Virtue Ethics & Utilitarianism & Deontology & Libertarianism \\
    Mistral-Large-3 & Utilitarianism & Virtue Ethics & Care Ethics & Deontology & Libertarianism \\
    o3-mini & Care Ethics & Virtue Ethics & Deontology & Utilitarianism & Libertarianism \\
    GPT-OSS-120B & Care Ethics & Utilitarianism & Virtue Ethics & Deontology & Libertarianism \\
    Phi-4 & Utilitarianism & Virtue Ethics & Care Ethics & Deontology & Libertarianism \\
    DeepSeek-V3 & Virtue Ethics & Utilitarianism & Care Ethics & Deontology & Libertarianism \\
    \bottomrule
  \end{tabular}
\end{table}

\begin{table}[ht]
  \centering
  \small
  \caption{Agreement between framework rankings obtained through direct ranking tasks and rankings reconstructed from morality scores. Agreement is measured using Spearman's $\rho$ and Kendall's $\tau$; values in parentheses are exact two-sided permutation $p$-values over the 120 possible orderings of the five frameworks. Higher values indicate stronger consistency across reporting mechanisms. The $p$-values test the significance of each model-level correlation across the five frameworks and are distinct from the group-level Mann--Whitney tests reported in Section~\ref{sec:reasoning}.}
  \label{tab:agreement}
  \begin{tabular}{lcc}
    \toprule
    Model & Spearman's $\rho$ ($p$) & Kendall's $\tau$ ($p$) \\
    \midrule
    DeepSeek-R1 & 1.0 (.017) & 1.0 (.017) \\
    DeepSeek-V3.2 & 0.7 (.233) & 0.6 (.233) \\
    GPT-4.1 & 0.8 (.133) & 0.6 (.233) \\
    GPT-4o & 0.6 (.350) & 0.4 (.483) \\
    GPT-5.2 (high) & 1.0 (.017) & 1.0 (.017) \\
    GPT-5.2 (none) & 0.6 (.350) & 0.4 (.483) \\
    GPT-5.4 (none) & 0.8 (.133) & 0.6 (.233) \\
    GPT-5.4 (xhigh) & 0.9 (.083) & 0.8 (.083) \\
    GPT-5 (high) & 0.9 (.083) & 0.8 (.083) \\
    Grok-4.1-Fast & 0.7 (.233) & 0.6 (.233) \\
    Kimi-K2.5 & 1.0 (.017) & 1.0 (.017) \\
    Mistral-Large-3 & 0.6 (.350) & 0.4 (.483) \\
    o3-mini & 1.0 (.017) & 1.0 (.017) \\
    GPT-OSS-120B & 0.9 (.083) & 0.8 (.083) \\
    Phi-4 & 1.0 (.017) & 1.0 (.017) \\
    DeepSeek-V3 & 0.7 (.233) & 0.6 (.233) \\
    \bottomrule
  \end{tabular}
\end{table}